\pdfoutput=1
\documentclass[11pt]{article}

\usepackage{ACL2023}

\usepackage{times}
\usepackage{latexsym}

\usepackage[T1]{fontenc}
\usepackage[utf8]{inputenc}

\usepackage{microtype}

\usepackage{inconsolata}
\usepackage{paralist}
\usepackage{graphicx}
\usepackage{multirow}
\usepackage{tabularray}
\usepackage[normalem]{ulem}
\usepackage{xcolor}
\usepackage{colortbl}
\RequirePackage{adjustbox}
\RequirePackage{soul}

\definecolor{LightRed}{rgb}{1,0.50,0.50}
\definecolor{LightGreen}{rgb}{0.66, 0.89, 0.63}
\definecolor{LightOrange}{rgb}{1,0.84,0.80}

\definecolor{LightBlue}{rgb}{0.90,0.98,1}
\definecolor{LightOrange}{rgb}{1,0.84,0.80}
\definecolor{LightRed}{rgb}{1,0.50,0.50}
\definecolor{LightGrey}{rgb}{0.9,0.9,0.9}
\definecolor{Green}{rgb}{0.55,0.70,0.0}
\definecolor{etonblue}{rgb}{0.59, 0.78, 0.64}
\definecolor{grannysmithapple}{rgb}{0.66, 0.89, 0.63}
\definecolor{palespringbud}{rgb}{0.93, 0.92, 0.74}
\definecolor{lightgray}{rgb}{0.83, 0.83, 0.83}
\definecolor{teagreen}{rgb}{0.82, 0.94, 0.75}
\definecolor{LightGreen1}{rgb}{0.59, 0.78, 0.64}
\definecolor{LightGreen2}{rgb}{0.66, 0.89, 0.63}
\definecolor{LightGreen3}{rgb}{0.82, 0.94, 0.75}
\definecolor{LightGreen4}{rgb}{0.93, 0.92, 0.74}
\definecolor{LightYellow}{rgb}{0.94,0.98,0.85}

\newcommand{\red}[1]{\textcolor{red}{#1}}

\newcommand{\am}[1]{\textcolor{red}{[AM: #1]}}

\def\hlc#1#2{\sethlcolor{#1}\hl{#2}}

\title{Low-Resource Counterspeech Generation for Indic Languages:\\ The Case of Bengali and Hindi}

\author{Mithun Das$^*$ \hspace{20pt} {\bf Saurabh Kumar Pandey$^*$} \hspace{20pt}{\bf Shivansh Sethi} \\ 
        {\bf Punyajoy Saha} \hspace{20pt} {\bf Animesh Mukherjee} \\
        Indian Insitute of Technology Kharagpur, India \\ 
         \texttt{mithundas@iitkgp.ac.in}, \{saurabh2000.iitkgp, shivanshsethi8821\}@gmail.com  \\
        \texttt{punyajoys@iitkgp.ac.in, animeshm@cse.iitkgp.ac.in}}
        
\newcommand\blfootnote[1]{%
  \begingroup
  \renewcommand\thefootnote{}\footnote{#1}%
  \addtocounter{footnote}{-1}%
  \endgroup
}

\begin{document}
\maketitle
\begin{abstract}
With the rise of online abuse, the NLP community has begun investigating the use of neural architectures to generate \textit{counterspeech} that can ``counter'' the vicious tone of such abusive speech and dilute/ameliorate their rippling effect over the social network. However, most of the efforts so far have been primarily focused on English. To bridge the gap for low-resource languages such as Bengali and Hindi, we create a benchmark dataset of 5,062 abusive speech/counterspeech pairs, of which 2,460 pairs are in Bengali, and 2,602 pairs are in Hindi. We implement several baseline models considering various interlingual transfer mechanisms with different configurations to generate suitable counterspeech to set up an effective benchmark\footnote{The benchmark dataset and source codes are available at \url{https://github.com/hate-alert/IndicCounterSpeech}}. We observe that the monolingual setup yields the best performance. Further, using synthetic transfer, language models can generate counterspeech to some extent; specifically, we notice that transferability is better when languages belong to the same language family. \red{\textit{Warning: Contains potentially offensive language.}}
\end{abstract}
\blfootnote{\textsuperscript{\bf{*}}Equal Contribution}

\section{Introduction}

The rise of online hostility has become an ominous issue endangering the safety of targeted people and groups and the welfare of society as a whole~\cite{statt2017youtube,vedeler2019hate,johnson2019hidden}. Therefore, to mitigate the widespread use of such hateful content, social media platforms generally rely on content moderation, ranging from deletion of hostile posts, shadow banning, suspension of the user account, etc.~\cite{tekirouglu2022using}. However, these strategies could impose restrictions on freedom of expression~\cite{myers2018censored}. Hence one of the alternative approaches to combat the rise of such hateful content is counterspeech (CS). CS is defined as a non-negative direct response to abusive speech (AS) that strives to denounce it by diluting its effect while respecting human rights. 

It has already been observed that many NGOs are deploying volunteers to respond to such hateful posts to keep the online space healthy~\cite{chung2019conan}. Even social media platforms like Facebook have developed guidelines for the general public to counter abusive speech online\footnote{\url{https://counterspeech.fb.com/en/}}. However, due to the sheer volume of abusive content,  it is an ambitious attempt to manually intervene all hateful posts. Thus, a line of NLP research focuses on semi or fully-automated generation models to assist volunteers involved in writing counterspeech~\cite{tekirouglu2020generating,chung2020italian,fanton2021human,zhu2021generate}. These generation models seek to minimize human intervention by providing ideas to the counter speakers that they can further post-edit if required.

However, the majority of these studies are concentrated on the English language. Hence effort is needed to develop datasets and language models (LMs) for low-resource languages. In the past few years, several smearing incidents, such as online anti-religious propaganda, cyber harassment, smearing movements, etc., have been observed in Bangladesh and India~\cite{das2022data}. Bangladesh has more than 150 million people with Bengali as the official language\footnote{\url{https://en.wikipedia.org/wiki/Bangladesh}}, and India has more than 1.3 billion people, with Hindi and English as the official language\footnote{\url{https://en.wikipedia.org/wiki/India}}. So far, several works have been done to detect malicious content in Bengali and Hindi~\cite{mandl2019overview,das2022hate}. However, no work has been done to generate automatic counterspeech for these languages.

Our key contributions in this paper are as follows:
\begin{compactitem}
    \item To bridge the research gap, in this paper, we develop a benchmark dataset of 5,062 AS-CS pairs, of which 2,460 pairs are in Bengali and 2,602 pairs are in Hindi. We further label the type of CS being used~\cite{benesch2016counterspeech}.
    \item We experiment with several transformer-based baseline models for CS generation considering GPT2, MT5,  BLOOM, ChatGPT, etc. and evaluate several interlingual mechanisms. 
    \item We observe that overall the monolingual setting yields the best performance across all the setups. Further, we notice that transfer schemes are more effective when languages belong to the same language family.
\end{compactitem}

\section{Related works}
This section briefly discusses the relevant work for abusive speech countering on social media platforms and the existing methodologies for CS generation strategies.

\noindent\textit{Online abuse countering}: A series of works have investigated online abusive content, aiming to study the online diffusion of abuse~\cite{mathew2019spread} and creating datasets for abuse detection~\cite{davidson2017automated,mandl2019overview,das2022hate} considering several multilingual languages. In many cases such detection models are used to censor abusive content which may curb the freedom of speech~\cite{myers2018censored}. Therefore as an alternative, NGOs have started employing volunteers to counter online abuse~\cite{chung2019conan}. Previous studies on countering abusive speech cover several aspects of CS, including defining counterspeech~\cite{benesch2016considerations}, studying their effectiveness~\cite{wright2017vectors}, and linguistically characterizing online counter speakers' accounts~\cite{mathew2019thou}.\\
\noindent\textit{CS dataset}: So far, several strategies have been followed for the collection of counterspeech datasets. \citet{mathew2019thou} crawled comments from Youtube with the replies to that comments and manually annotated the hateful posts along with the counterspeech responses. \citet{chung2019conan} created three multilingual datasets in English, French, and Italian. To construct the dataset, the authors asked native expert annotators to write hate speech, and with the effort of more than 100 operators from three different NGOs, they built the overall dataset. \citet{fanton2021human} proposed a novel human-in-the-loop data collection process in which a generative language model is refined iteratively. To our knowledge, no dataset has been built for low-resource languages such as Bengali and Hindi; therefore, in this work, we construct a new benchmark dataset of 5,062 AS-CS pairs for two Indic languages -- Bengali and Hindi.\\
\noindent\textit{CS generation}: Several studies have been conducted for the generation of effective counterspeech. \citet{qian2019benchmark} employ a mix of automatic and human interventions to generate counternarratives. \citet{tekirouglu2020generating} presented novel techniques to generate counterspeech using a GPT-2 model with post-facto editing by the experts/annotator groups. Zhu and Bhat~\cite{zhu2021generate} suggested an automated pipeline of candidate CS generation and filtering.  \citet{chung2020italian} investigated the generation of Italian CS to fight online hate speech. Recently \citet{tekirouglu2022using} performed a comparative study of counter-narratives generations considering several transformer-based models such as GPT-2, T5, etc. So far, no work has examined the generation of counterspeech for under-resourced languages such as Bengali and Hindi; therefore, we attempt to fill this critical gap by benchmarking various transformer-based language models.

\section{Dataset creation}
\subsection{Seed sets}
\noindent\textbf{Data collection \& sampling}: To create the CS dataset, we need a seed set of abusive posts for which the counterspeech could be written. For this purpose, we first create a set of abusive lexicons for Bengali and Hindi. We search for tweets using the Twitter API containing phrases from the lexicons, resulting in a sample of 100K tweets for Bengali and 200K for Hindi. The presence of an abusive lexicon in a post does not ensure that the post is abusive; therefore, we randomly sample around 3K data points from both languages and annotate the sample dataset to find out the abusive tweets.

\noindent\textbf{Annotation:} We define a post as abusive if it dehumanizes or incites harm towards an individual or a community. It can be done using derogatory or racial slur words within the post targeting a person based on protected attributes such as race, religion, ethnic origin, sexual orientation, disability, or gender~\cite{gupta2022multilingual}. Based on the defined guidelines, two PhD students annotated the posts as abusive or non-abusive. Both students have extensive prior experience working with malicious content on social media. After completing the annotation, we remove the conflicting cases and keep the posts labeled as abusive by both annotators. To measure the annotation quality, we compute the inter-annotator agreement achieving a Cohen's $\kappa$ of 0.799. Additionally, to increase the diversity of abusive speech in the dataset, we randomly select some annotated abusive speech data points from existing annotated datasets for both Bengali~\cite{das2022hate} and Hindi~\cite{mandl2019overview}.

\subsection{Guidelines for writing counterspeech}
Before writing the counterspeech, we develop a set of guidelines that the annotators have to follow to make the writing effective. We define counterspeech as any direct response to abusive or hateful speech which seeks to undermine it without harassing or using an aggressive tone towards the hateful speaker. There could be several techniques to counter abusive speech. \citet{benesch2016considerations} defines eight strategies that speakers typically use to counter abusive speech. However, not all of these strategies effectively reduce the propagation of abusive speech. A counterspeech can be deemed successful if it has a positive impact on the hateful speaker. Therefore, the authors further recommended strategies that can facilitate positive influence. As a result, we instructed the annotators to follow the following strategies: \textit{warning of consequences}, \textit{pointing out hypocrisy}, \textit{shaming \& labeling}, \textit{affiliation}, \textit{empathy}, and \textit{humor \& sarcasm} (see Appendix \ref{sec:anngd} for more details).

\noindent{\textbf{Annotation process}}:
We use the Amazon Mechanical Turk (AMT) developer sandbox for our annotation task. For the annotation process, we hire 11 annotators, including undergraduate students and researchers in NLP: seven were males, four were females, and all were 24 to 30 years old. Among the 11 annotators, seven are native Hindi speakers, and four are native Bengali speakers. We have given them three Indian rupees as compensation for writing each counterspeech, which is higher than the minimum wage in India~\cite{indiabriefingGuideMinimum}. Two expert PhD students with more than three years of experience in research in this area led the overall annotation process.

\begin{figure}[h]
  \centering
\includegraphics[width=0.68\linewidth]{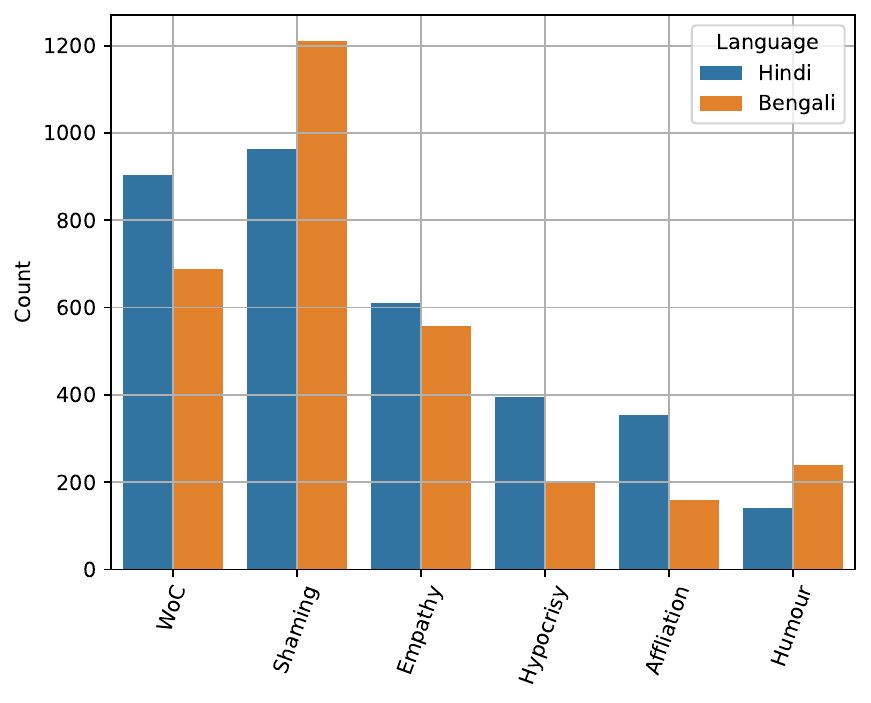}
\caption{\footnotesize Distribution of the different  types of CS based on human annotations.}
 \label{fig:strat}
\end{figure}

\subsection{Dataset Creation Steps}

Before starting with the actual annotation, we need a gold-label dataset to train the annotators. Initially, we wrote 20 counterspeech per language, which have been used to train the annotators. We schedule several meetings with the annotators to make them understand the guidelines and the drafted examples.

\noindent\textbf{Pilot annotation}: We conduct a pilot annotation on a subset of 10 abusive speech, which helped the annotators understand the counterspeech writing process task. We instruct the annotators to write counterspeech for an abusive speech according to the annotation guidelines. We told them to keep the annotation guidelines open in front of them while writing the counterspeech to have better clarity about the writing strategies. After the pilot annotation, we went through the counterspeech writings and manually checked to verify the annotators' understanding of the task. We observe that although the written counterspeech is appropriate, sometimes, the annotators mislabel the strategy. We consult with them regarding their incorrect strategy labeling so that they could rectify them while doing the subsequent annotations. The pilot annotation is a crucial stage for any dataset creation process as these activities help the annotators better understand the task by correcting their mistakes. In addition, we collect feedback from annotators to enrich the main annotation task.

\begin{table}[]
\small
\begin{tabular}{l|cc|cc|}
\cline{2-5}
 & \multicolumn{2}{c|}{\textbf{Diversity}} & \multicolumn{2}{c|}{\textbf{Novelty}} \\ \hline
\multicolumn{1}{|l|}{\textbf{Dataset}} & \multicolumn{1}{c|}{AS} & CS & \multicolumn{1}{c|}{AS} & CS \\ \hline
\multicolumn{1}{|l|}{CONAN} & \multicolumn{1}{c|}{0.5245} & 0.7215 & \multicolumn{1}{c|}{0.9108} & 0.9237 \\ \hline
\multicolumn{1}{|l|}{Bengali (Ours)} & \multicolumn{1}{c|}{0.8172} & 0.6979 & \multicolumn{1}{c|}{0.9868} & 0.9553 \\ \hline
\multicolumn{1}{|l|}{Hindi (Ours)} & \multicolumn{1}{c|}{0.7745} & 0.6640 & \multicolumn{1}{c|}{0.9616} & 0.9089 \\ \hline
\multicolumn{1}{|l|}{Total (Ours)} & \multicolumn{1}{c|}{0.7953} & 0.6805 & \multicolumn{1}{c|}{0.9742} & 0.9321 \\ \hline
\end{tabular}
\caption{Diversity and novelty scores of AS and CS for our proposed datasets and their comparison with the CONAN~\cite{fanton2021human} dataset.}
\label{tab:datastats}
\end{table}

\noindent\textbf{Main annotation:} After the pilot annotation stage, we proceed with the main annotation task. We gave them 20 abusive speech posts per week for writing the counterspeech. Since consuming a lot of abusive content can have a negative psychological impact on the annotators, we kept the timeline relaxed and suggested they take at least 5 minute break after writing each counterspeech. Finally, we also had regular meetings with them to ensure that they did not have any adverse effects on their mental health. Our final dataset consists of 5,062 AS-CS pairs, of which 2,460 pairs are in Bengali and 2,602 pairs are in Hindi. We assess the quality of the generated dataset based on the diversity and novelty metrics; the results are noted in Table \ref{tab:datastats}. The scores are considerably better than the existing CONAN counterspeech dataset which is a de facto benchmark in the literature~\cite{fanton2021human} in English. Further we illustrate the distribution of different types of CS in Figure~\ref{fig:strat}.

\if{0}\begin{table*}
\begin{tabular}{ll}
\centering
%\begin{minipage}[0.5\textwidth]

\includegraphics[width=0.3\textwidth]{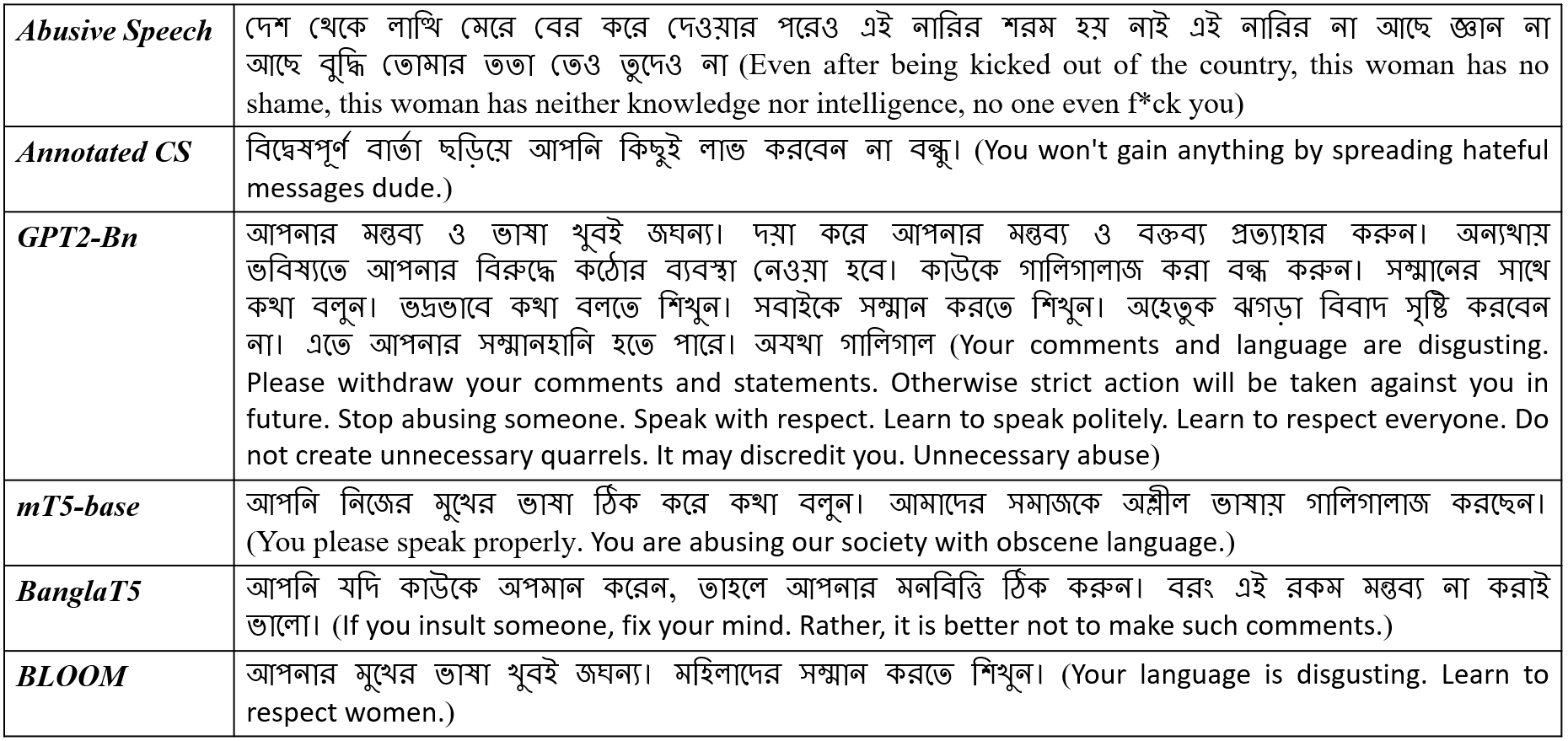}
%\caption{Examples of AS-CS pairs generated by some of the models for the Bengali language. (monolingual setting) \am{Add chatgpt}}
% \label{conter1}
 %\end{minipage}
 &
 %\begin{minipage}[0.5\textwidth]
%\centering
\includegraphics[width=0.3\textwidth]{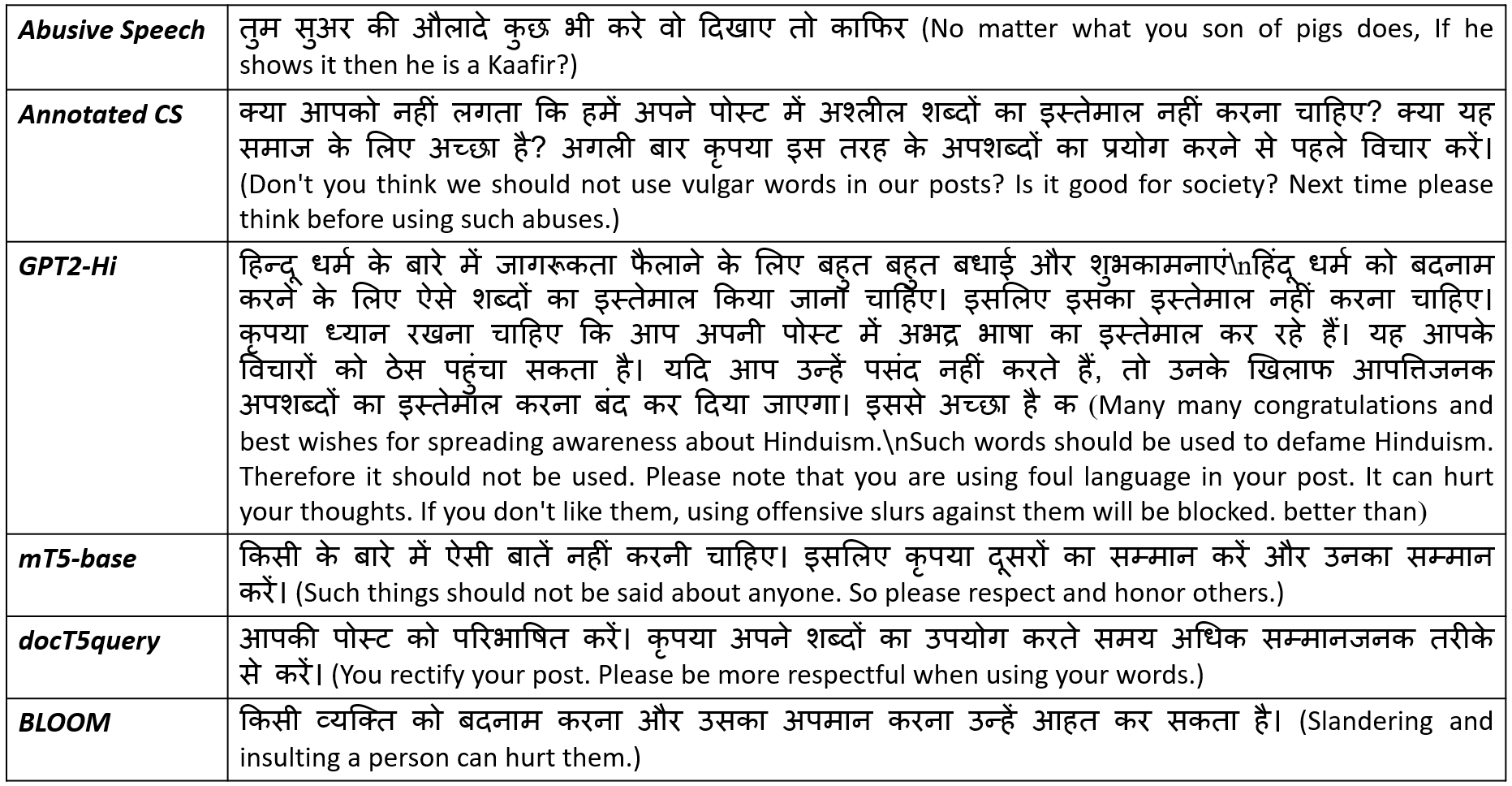}
 \caption{\footnotesize Examples of AS-CS pairs generated by some of the models for the Hindi language. (monolingual setting) \am{Add chatgpt}}
\label{conter2}
%\end{minipage}
\end{tabular}
\end{table*}\fi

\section{Methodology}
\subsection{Baseline models}
In this section, we discuss the models we implement for the automatic generation of counterspeech. We experiment with a wide range of models.

\noindent\textbf{GPT-2}: GPT-2~\cite{radford2019language} is an unsupervised generative model released by OpenAI only supports the English language. Our focus is to generate counterspeech for non-English language. Therefore to generate counterspeech for Hindi, we use the \textbf{GPT2-Hindi} (GPT2-HI)~\cite{ParmarHindi}  model, and for Bengali, we use the \textbf{GPT2-bengali} (GPT2-BN)~\cite{CommunityBengali} model published on Huggingface~\cite{wolf2019huggingface}.

\noindent\textbf{T5-based models}: mT5~\cite{xue2020mt5}, a multilingual variant of T5, is an encoder-decoder model pre-trained on 101 languages released by Google. The mT5 model has five variants, and we use the mT5-base variant for our experiments. For the Hindi language, we also use a fine-tuned mT5-base model,  \texttt{docT5query-Hindi}~\cite{nogueira2019document}, which is trained on a (query passage) from the mMARCO dataset. For Bengali, we also experiment with the \texttt{BanglaT5}~\cite{bhattacharjee2022banglanlg} model, which is pre-trained with a clean corpus of 27.5 GB Bengali data.

\noindent\textbf{BLOOM}: BLOOM~\cite{scao2022bloom}  is an autoregressive large language model developed to continue text from a prompt utilizing highly efficient computational resources on vast amounts of text data, can be trained to accomplish text tasks it has not been explicitly instructed for by casting them as text generation tasks.

\noindent\textbf{ChatGPT}: ChatGPT~\cite{openaichatGPT} is a robust large language model developed by OpenAI, capable of performing various natural language processing tasks such as question answering, language translation, text completion, and many more.

\subsection{Interlingual transfer mechanisms}

We perform three sets of experiments to check how different models perform under various settings. Specially, we investigate the benefits of using  silver label counterspeech datasets to improve the performance of the language models for better counterspeech generation. Below we illustrate the details of these experiments\footnote{For ChatGPT, we only generate CSs in a zero-shot setting. We refrained from fine-tuning due to budget constraints and high computational resource requirements, making it impractical to conduct such experiments.}.

\noindent\textbf{Monolingual setting}: In this setting, we use the same language's gold data points for training, validation, and testing for the counterspeech generation. This scenario generally emerges in the real world, where monolingual datasets are developed and utilized to create classification models, generation models, or models for any other downstream task. Simulating this scenario is more expensive as the gold label dataset has to be built from scratch. In our case, it is the AS-CS dataset.

\noindent\textbf{Joint training}: In this setup, while training a model, we combine the datasets of both the Bengali and Hindi languages. The idea is, even though the characters and words used to represent different languages vary, how will these language generation models perform if one wants to create a generalizable model to handle counterspeech generation for multiple languages?

\noindent\textbf{Synthetic transfer}: Due to the less availability of datasets in low-resource languages, in this strategy, we experiment whether resource-rich languages can be helpful if we translate them into low-resource languages and build the generation model from scratch. Further, we experiment that even if some low-resource language datasets are available belonging to the same language community, will it be helpful to generate suitable counterspeeches for other languages? To accomplish this, we use one of the experts annotated English CS datasets~\cite{fanton2021human} (typically constructed with a human-in-the-loop) and translate it into Hindi and Bengali to develop synthetic (silver) counterspeech datasets. Also, we translate the Bengali AS-CS pairs to Hindi and vice-versa to check language transferability between the same language community. In summary, we create the following four synthetic datasets: \textbf{EN $\rightarrow$ BN}, \textbf{HI $\rightarrow$ BN}, \textbf{EN $\rightarrow$ HI}, and \textbf{BN $\rightarrow$ HI}\footnote{Languages are represented by ISO 639-1 codes.}. We use Google Translate API\footnote{\url{https://cloud.google.com/translate}} to perform the translation. Next using the synthetic counterspeech dataset, we build our generation model. In the zero-shot setting (\textbf{STx0}), we do not use any gold target instances. In a related few-shot setting, we allow $n$ = 100 and 200 pairs from the available gold AS-CS pairs to fine-tune the generation models. These are called \textbf{STx1} and \textbf{STx2}.

\begin{table*}
\scriptsize
\centering
\begin{tabular}{|l|cccc|c|c|c|c|cccc|} 
\hline
\multicolumn{13}{|c|}{\textbf{Bengali}}   \\ 
\hline
\multirow{2}{*}{\textbf{Model}} & \multicolumn{4}{c|}{\textbf{Overlap}}                                                  & \multicolumn{1}{l|}{\textbf{BERT SC}} & \multicolumn{1}{c}{\textbf{Diversity}} & \textbf{Novelty} & \textbf{Abuse} & \multicolumn{4}{c|}{\textbf{Human evaluation}}                 \\ 
\cline{2-13}
                                & \textbf{B-2}    & \textbf{B-3}                       & \textbf{M}     & \textbf{ROU}   & \multicolumn{1}{l|}{}                 & \textbf{-}                             & \textbf{-}       & \textbf{-}     & \textbf{SUI}  & \textbf{SPE}  & \textbf{GRM}  & \textbf{CHO}   \\ 
\hline
GPT2-BN                         & 0.053           & 0.039                              & \uline{0.098}  & 0.166          & 0.665                                 & \underline{0.598}                         & \underline{0.807}   & 0.856              & 3.07          & 2.75          & 3.47          & 0.74           \\
mT5-base                        & \uline{0.117}   & \uline{0.099}                      & 0.093          & \uline{0.178}  & \uline{0.731}                         & 0.314                                  & 0.637            & 0.964              & 3.65          & \uline{3.07}  & \uline{4.03}  & \textbf{0.90}  \\
BanglaT5                        & \textbf{0.130}  & \textbf{0.102}                     & \textbf{0.119} & \textbf{0.209} & 0.724                                 & 0.549                          & 0.714    & \uline{0.972}              & \textbf{3.74} & \textbf{3.15} & 3.77          & \uline{0.88}   \\
BLOOM                           & 0.093           & 0.084                              & 0.067          & 0.139          & \textbf{0.732}                        & 0.014                                  & 0.567            & \textbf{0.991}             & \uline{3.73}  & 3.05          & \textbf{4.42} & \textbf{0.90}  \\ 
ChatGPT                        & 0.024           & 0.019                              & 0.069          & 0.094          & 0.661                        & \textbf{0.850}                                  & \textbf{0.914}            & 0.746             & 2.58  & 2.44          & 3.83 & 0.615  \\ 
\hline
\multicolumn{13}{|c|}{\textbf{Hindi}}                                                                                                                                                                                                                                                                          \\ 
\hline
GPT2-HI                         & 0.101           & \multicolumn{1}{l}{0.067}          & \underline{0.140} & 0.244  & 0.651                                 & \underline{0.510}                         & \underline{0.778}   & 0.641              & 2.96          & 3.12          & 3.10          & 0.72           \\
mT5-base                        & \textbf{ 0.175} & \multicolumn{1}{l}{\textbf{0.123}} & 0.133  & \underline{0.245} & \textbf{0.715}                        & 0.365                                  & 0.674            & \uline{0.902}              & \uline{3.47}  & \uline{3.15}  & \uline{4.26}  & \uline{0.92}   \\
docT5query                      & 0.140           & \multicolumn{1}{l}{0.103}          & 0.110          & 0.221          & 0.698                                 & 0.399                          & 0.774    & 0.608              & 2.75          & 2.43          & 4.16          & 0.60           \\
BLOOM                           & \uline{0.145}   & \multicolumn{1}{l}{\uline{0.108}}  & 0.103          & 0.202          & \uline{0.712}                         & 0.064                                  & 0.637            & \textbf{0.917}             & \textbf{3.58} & \textbf{3.16} & \textbf{4.69} & \textbf{0.94}  \\
ChatGPT                           & 0.070           & 0.040                              & \textbf{0.166}          & \textbf{0.261}          & 0.673                        & \textbf{0.752}                                  & \textbf{0.820}            & 0.743             & 2.08  & 2.48          & 4.04 & 0.54  \\ 
\hline
\end{tabular}
\caption{Quantitative results of fine-tuned models (monolingual setting) . BERT SC: BERTScore, docT5query: docT5query-Hindi.}
\label{tab1F}
\end{table*}

\begin{table*}
\scriptsize
\centering
\begin{tabular}{|l|cccc|c|c|c|c|cccc|} 
\hline
\multicolumn{13}{|c|}{\textbf{Bengali}}                                                                                                                                                                                                                                                                    \\ 
\hline
\multirow{2}{*}{\textbf{Model}} & \multicolumn{4}{c|}{\textbf{Overlap}}                                                  & \multicolumn{1}{l|}{\textbf{BERT SC}} & \multicolumn{1}{c}{\textbf{Diversity}} & \textbf{Novelty} & \textbf{Abuse} & \multicolumn{4}{c|}{\textbf{Human evaluation}}             \\ 
\cline{2-13}
                                & \textbf{B-2}    & \textbf{B-3}                       & \textbf{M}     & \textbf{ROU}   & \multicolumn{1}{l|}{}                 & \textbf{-}                             & \textbf{-}       & \textbf{-}     & \textbf{SUI} & \textbf{SPE} & \textbf{GRM} & \textbf{CHO}  \\ 
\hline
mT5-base                        & \textbf{0.101}  & \textbf{0.087}                     & \textbf{0.076} & 0.150          & 0.718                                 & \textbf{0.401}                         & \textbf{0.692}   & 0.967              & 3.14           & \textbf{2.71}           & 4.25           & 0.85   \\
BLOOM                           & 0.078           & 0.071                              & 0.070          & \textbf{0.167} & \textbf{0.727}                        & 0.033                                  & 0.597            & \textbf{0.980}              & \textbf{3.25}           & 2.67           & \textbf{4.82}  & \textbf{0.91}   \\ 
\hline
\multicolumn{13}{|c|}{\textbf{Hindi}}                                                                                                                                                                                                                                                                      \\ 
\hline
mT5-base                        & \textbf{ 0.174} & \multicolumn{1}{l}{\textbf{0.125}} & \textbf{0.129} & \textbf{0.238} & 0.713                                 & \textbf{0.391}                         & \textbf{0.695}   & 0.893              & \textbf{3.38}   & \textbf{3.28}   & \textbf{4.34}   & 0.80            \\
BLOOM                           & 0.089           & \multicolumn{1}{l}{0.076}          & 0.073          & 0.161          & \textbf{0.717}                        & 0.007                                  & 0.593            & \textbf{0.945}              & 2.99           & 2.73           & 3.94           & \textbf{0.95}            \\
\hline
\end{tabular}
\caption{\footnotesize Quantitative results of the fine-tuned models (joint training). BERT SC: BERTScore.}
\label{tab2f}
\end{table*}

\subsection{Experimental setup}
This section describes the training and evaluation approach followed for the language generation models.

\subsubsection{Training} All models except ChatGPT were evaluated using the same 70:10:20 train, validation, and test split, ensuring no repetition of AS across sets. For the synthetic transfer learning experiments, we split the synthetic datasets into an 85:15 train-validation split. The test set remains exactly the same 20\% held out split as earlier. We use 100 and 200 AS-CS gold pairs to further fine-tune the model for the few-shot transfer learning experiments. We make three different random sets for each target dataset to make our evaluation more effective and report the average performance.

\begin{table}[h]
  \centering
\includegraphics[width=\linewidth]{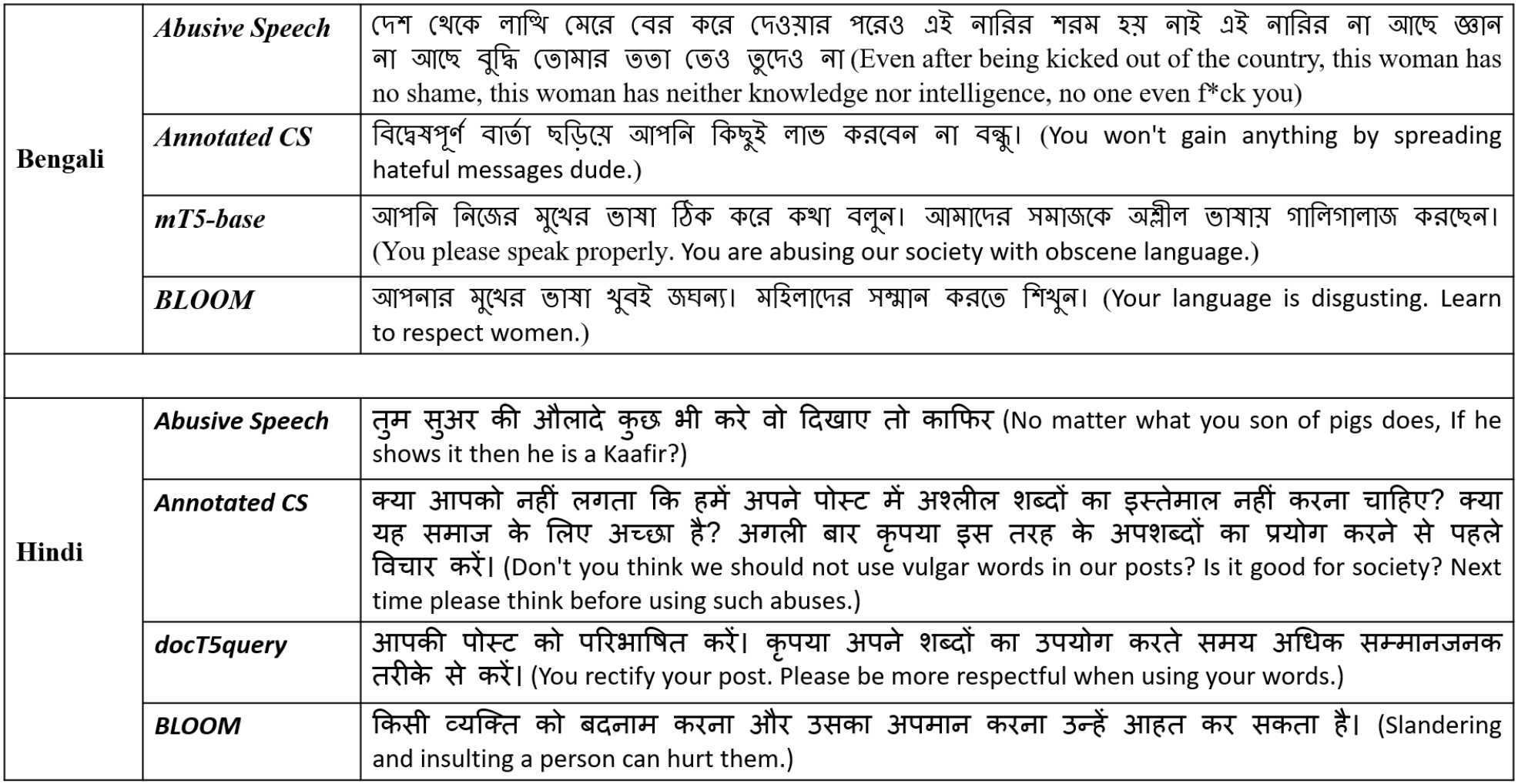}
\captionof{table}{\footnotesize  Examples of AS-CS pairs generated by some of the models (monolingual setting).}
 \label{tab:counterEx}
\end{table}

We use a simple regex-based preprocessing pipeline to remove special characters, URLs, emojis, etc. We limit the maximum length of AS-CS pairs to 400 to include both long and short texts. For the GPT-based and BLOOM models, we follow an autoregressive text generation approach where we separate AS and CS pairs by ‘EOS BOS’ token to guide the generation to predict suitable CS. For the T5-based models, we use the ‘counterspeech’ token as the prompt for input and annotated counterspeech as output (more details in Appendix \ref{imple_det}). For ChatGPT, our approach to addressing the specific problem of generating counter-speech for abusive language involves crafting well-designed prompts; we aim to generate counter-speech responses for a given abusive speech. We structure the prompts as follows: ``\textit{Please write a counter speech in <language name> for the provided abusive speech in <language name>: abusive speech}''. Using this prompt, we generate CSs for the test set that was used in all the other models.

\subsubsection{CS generation} Following previous research~\cite{tekirouglu2022using}, in our experiments, we use the following parameters as default: beam search with five beams and repetition penalty = 2; top-$k$ with $k = 40$;  top-$p$ with $p = .92$; min\_length = 20 and max\_length = 300.  We also use sampling to get more diverse generations.  We did not need to use any of these parameters for the ChatGPT model. Instead, we passed only the prompt and the AS for which CS had to be generated. We show examples of some generated CSs in Table \ref{tab:counterEx}.

\subsection{Evaluation metric}
We consider several metrics to evaluate various aspects of counterspeech generation. For all metrics, higher is better and the best performance in each column is marked in \textbf{bold}, and the second best is \underline{underlined}.

\noindent\textbf{Overlap metrics}: These metrics evaluate the quality of the generation model by comparing the $n$-gram similarity of the generated outputs to a set of reference texts. We use the counterspeech produced by the various models as candidates and our human written counterspeech as ground truths. To measure how closely the generated counterspeech resembles the ground truth counterspeech, we specifically employ BLEU (\textbf{B-2}, \textbf{B-3}), METEOR(\textbf{M}), and ROUGE-1 (\textbf{ROU}).

\noindent\textbf{Diversity metrics}: They are used to measure if the generation model produces diverse and novel counterspeech. We employ Jaccard similarity to compute the amount of novel content present in the generated CS compared to the ground truth.

\noindent\textbf{Abusiveness}: Finally, to measure the abusiveness of a text, we use \texttt{indic-abusive-allInOne-MuRIL} model~\cite{das2022data} trained on eight different Indic languages in two classes – abusive and non-abusive. We report the confidence between 0-1 for the non-abusive class.

\noindent\textbf{BERTScore}: It is an automatic evaluation metric for text generation. Analogously to common metrics, BERTScore~\cite{bert-score} computes a similarity score for each token in the candidate sentence with each token in the reference sentence. However, instead of exact matches, we compute token similarity using contextual embeddings. BERTScore correlates better with human judgments and provides stronger model selection performance than existing metrics. We compute BERTscore initialized with the \textit{bert-base-multilingual-cased} model~\cite{kenton2019bert}.

\noindent\textbf{Human evaluation metrics}: Despite being difficult to collect, human assessments furnish a more accurate evaluation and a deeper understanding than automatic metrics. Following the previous studies~\cite{chung2020italian,tekirouglu2022using}, we also conduct a human evaluation to compare the generation quality of the models under various settings. We use the following aspects for the assessment of generated counterspeech. \textbf{Suitableness} (SUI) measures how suitable the generated CS is in response to the input AS in terms of semantic relatedness and guidelines. \textbf{Specificity} (SPE) measures how specific are the explanations obtained by the generated CS as a response to the input AS. \textbf{Grammaticality} (GRM) measures how grammatically accurate the generated CS is. \textbf{Choose-or-not}(CHO) assesses if the annotators would choose that CS for post-editing and use in a real-life scenario as in the setup suggested by \citet{chung2021empowering}. \\
To perform the human evaluation, for each model, we randomly select 50 random AS-CS instances from the generated pairs and assign our trained annotators to check the generated CS quality manually.

\begin{table*}
\scriptsize
\centering
\begin{tabular}{|l|cccc|c|c|c|c|cccc|} 
\hline
\multicolumn{13}{|c|}{\textbf{English -\textgreater{} Bengali}}                                                                                                                                                                                                                                               \\ 
\hline
\multirow{2}{*}{\textbf{Model}} & \multicolumn{4}{c|}{\textbf{Overlap}}                                                 & \multicolumn{1}{l|}{\textbf{BERT SC}} & \multicolumn{1}{c}{\textbf{Diversity}} & \textbf{Novelty} & \textbf{Abuse} & \multicolumn{4}{c|}{\textbf{Human evaluation}}                 \\ 
\cline{2-13}
                                & \textbf{B-2}   & \textbf{B-3}                       & \textbf{M}     & \textbf{ROU}   & \multicolumn{1}{l|}{}                 & \textbf{-}                             & \textbf{-}       & \textbf{-}     & \textbf{SUI}  & \textbf{SPE}  & \textbf{GRM}  & \textbf{CHO}   \\ 
\hline
GPT2-BN                         & 0.029          & 0.025                              & \textbf{0.044} & 0.094          & 0.623                                 & \textbf{0.725}                         & \textbf{0.899}   & 0.672              & 1.03          & 1.03          & 2.05          & 0.01           \\
mT5-base                        & \uline{0.064}  & \textbf{0.058}                     & 0.042          & \uline{0.095}  & \textbf{0.689}                        & 0.468                                  & 0.863            & 0.813              & \uline{1.16}  & \uline{1.13}  & \uline{2.42}  & \textbf{0.12}  \\
BanglaT5                        & \textbf{0.065} & \textbf{0.058}                     & \textbf{0.054} & \textbf{0.124} & \uline{0.676}                         & \uline{0.515}                          & \uline{0.870}    & \uline{0.828}              & 1.02          & 1.02          & 1.61          & 0.01           \\
BLOOM                           & 0.046          & \uline{0.043}                      & 0.030          & 0.078          & 0.658                                 & 0.210                                  & 0.865            & \textbf{0.976}              & \textbf{1.17} & \textbf{1.15} & \textbf{2.54} & \uline{0.10}   \\ 
\hline
\multicolumn{13}{|c|}{\textbf{Hindi -\textgreater{} Bengali}}                                                                                                                                                                                                                                                 \\ 
\hline
GPT2-BN                         & 0.026          & \multicolumn{1}{l}{0.020}          & \textbf{0.067} & \textbf{0.140} & 0.616                                 & \uline{0.522}                          & \textbf{0.852}   & 0.911              & \textbf{2.32} & \textbf{2.04} & \uline{3.03}  & \textbf{0.60}  \\
mT5-base                        & \uline{ 0.080} & \multicolumn{1}{l}{\textbf{0.072}} & 0.056          & 0.120          & \uline{0.702}                         & 0.346                                  & 0.815            & \uline{0.981}              & 2.17          & \uline{1.92}  & 3.07          & \uline{0.54}   \\
BanglaT5                        & \textbf{0.081} & \multicolumn{1}{l}{\uline{0.070}}  & \uline{0.064}  & \uline{0.136}  & 0.691                                 & \textbf{0.601}                         & \uline{0.838}    & 0.974              & 1.70          & 1.55          & 2.44          & 0.32           \\
BLOOM                           & 0.059          & \multicolumn{1}{l}{0.056}          & 0.037          & 0.089          & \textbf{0.705}                        & 0.027                                  & 0.825            & \textbf{0.988}              & \uline{2.09}  & 1.79          & \textbf{3.15} & 0.36           \\
\hline
\end{tabular}
\caption{\footnotesize Quantitative results of fine-tuned models for the zero-shot synthetic transfer for Bengali test set. BERT SC: BERTScore.}
\label{tab3}
\end{table*}

\begin{table*}
\scriptsize
\centering
\begin{tabular}{|l|cccc|c|c|c|c|cccc|} 
\hline
\multicolumn{13}{|c|}{\textbf{English -\textgreater{} Hindi}}                                                                                                                                                                                                                                                  \\ 
\hline
\multirow{2}{*}{\textbf{Model}} & \multicolumn{4}{c|}{\textbf{Overlap}}                                                  & \multicolumn{1}{l|}{\textbf{BERT SC}} & \multicolumn{1}{c}{\textbf{Diversity}} & \textbf{Novelty} & \textbf{Abuse} & \multicolumn{4}{c|}{\textbf{Human evaluation}}                 \\ 
\cline{2-13}
                                & \textbf{B-2}    & \textbf{B-3}                       & \textbf{M}     & \textbf{ROU}   & \multicolumn{1}{l|}{}                 & \textbf{-}                             & \textbf{-}       & \textbf{-}     & \textbf{SUI}  & \textbf{SPE}  & \textbf{GRM}  & \textbf{CHO}   \\ 
\hline
GPT2-HI                         & 0.073           & 0.049                              & \uline{0.106}  & \uline{0.217}  & 0.626                                 & \textbf{0.585}                         & \textbf{0.813}   & \uline{0.765}              & 1.11          & 1.09          & 2.17          & 0.06           \\
mT5-base                        & \textbf{0.142}  & \textbf{0.100}                     & \textbf{0.107} & \textbf{0.221} & \textbf{0.694}                        & \uline{0.501}                          & 0.779            & 0.700              & 1.25          & 1.20          & \uline{3.02}  & 0.16           \\
docT5Query                      & \uline{0.125}   & \uline{0.093}                      & 0.089          & 0.197          & \uline{0.689}                         & 0.462                                  & \uline{0.795}    & 0.589              & \uline{1.33}  & \textbf{1.29} & \textbf{3.09} & \textbf{0.23}  \\
BLOOM                           & 0.113           & 0.082                              & 0.092          & 0.209          & 0.679                                 & 0.307                                  & 0.778            & \textbf{0.794}              & \textbf{1.32} & \uline{1.26}  & 2.95          & \uline{0.17}   \\ 
\hline
\multicolumn{13}{|c|}{\textbf{Bengali -\textgreater{} Hindi}}                                                                                                                                                                                                                                                  \\ 
\hline
GPT2-HI                         & 0.082           & \multicolumn{1}{l}{0.055}          & \textbf{0.127} & \textbf{0.249} & 0.647                                 & \textbf{0.302}                         & \uline{0.786}    & \uline{0.827}              & 2.40          & 2.46          & 3.20          & 0.04           \\
mT5-base                        & \textbf{ 0.169} & \multicolumn{1}{l}{\textbf{0.121}} & \uline{0.123}  & \uline{0.228}  & \textbf{0.698}                        & \uline{0.179}                          & 0.742            & 0.564              & \uline{3.46}  & \uline{3.26}  & \uline{4.18}  & \uline{0.58}   \\
docT5Query                      & \uline{0.144}   & \multicolumn{1}{l}{\uline{0.107}}  & 0.101          & 0.196          & 0.693                                 & 0.123                                  & 0.769            & 0.530              & \textbf{3.86} & \textbf{3.56} & \textbf{4.60} & \textbf{0.82}  \\
BLOOM                           & 0.097           & \multicolumn{1}{l}{0.078}          & 0.067          & 0.159          & \uline{0.697}                         & 0.084                                  & \textbf{0.793}   & \textbf{0.860}              & 2.48          & 2.64          & 3.54          & 0.12           \\
\hline
\end{tabular}
\caption{\footnotesize Quantitative results of fine-tuned models for the zero-shot synthetic transfer for Hindi test set. BERT SC: BERTScore, docT5Query: docT5Query-Hindi.}
\label{tab4}
\end{table*}

\section{Results} 
%\md{The whole result section is re-written}
\subsection{Performance in the monolingual setting}
In Table \ref{tab1F}, we report the performance in the monolingual setting. We observe that -- \\
For the Bengali language, \texttt{BanglaT5} model performs the best across all the \textbf{overlapping metrics} (\textbf{B-2}: 0.130, \textbf{B-3}: 0.102, \textbf{M}: 0.119, \textbf{ROU}: 0.209), while the mT5-base model performs the second best in terms of \textbf{BLEU} \& \textbf{ROU} metrics. 
When considering \textbf{BERTScore}, we find that BLOOM achieves the highest score (0.732), closely followed by the mT5-base achieves the second-Highest score (0.731).
We notice that BLOOM exhibits the lowest performance in terms of \textbf{diversity} (0.014) and \textbf{novelty} (0.567), implying that it tends to produce similar responses. In contrast, we observe that ChatGPT exhibited the highest performance, while GPT2-BN exhibited the second-highest score. This indicates that the large language model ChatGPT can generate more diverse counterspeeches compared to the other models. 
All the models generate mostly non-abusive counterspeeches, with BLOOM achieving the highest score of 0.991 and  \texttt{BanglaT5} attaining the second-best score of 0.972.
In terms of human judgments, the  \texttt{BanglaT5} model achieves the highest score in terms of \textbf{suitableness} \& \textbf{specificity}. The mT5-base \& BLOOM models demonstrate superior performance in the \textbf{choose-or-not} metric.  In contrast, ChatGPT showed inferior performance in the \textbf{choose-or-not} metric, indicating that its responses were not as good to be chosen as counterspeeches in response to an abusive speech.

\begin{table}[]
\centering
\scriptsize
\begin{tabular}{|ccccccc|}
\hline
\multicolumn{7}{|c|}{\textbf{English -> Bengali}} \\ \hline
\multicolumn{1}{|l|}{} & \multicolumn{2}{c|}{\textbf{B-2}} & \multicolumn{2}{c|}{\textbf{M}} & \multicolumn{2}{c|}{\textbf{ROU}} \\ \hline
\multicolumn{1}{|l|}{\textbf{Model}} & \multicolumn{1}{l|}{STx1} & \multicolumn{1}{l|}{STx2} & \multicolumn{1}{l|}{STx1} & \multicolumn{1}{l|}{STx2} & \multicolumn{1}{l|}{STx1} & STx2 \\ \hline
\multicolumn{1}{|l|}{GPT2-BN} & \multicolumn{1}{l|}{\hlc{LightGreen3}{0.088}} & \multicolumn{1}{l|}{0.027} & \multicolumn{1}{l|}{\hlc{LightGreen3}{0.045}} & \multicolumn{1}{l|}{\hlc{LightGreen1}{0.057}} & \multicolumn{1}{l|}{\hlc{LightGreen3}{0.100}} & \hlc{LightGreen1}{0.122} \\ \hline
\multicolumn{1}{|l|}{mT5-base} & \multicolumn{1}{l|}{\hlc{LightGreen3}{0.107}} & \multicolumn{1}{l|}{\hlc{LightGreen1}{0.114}} & \multicolumn{1}{l|}{\hlc{LightGreen3}{0.079}} & \multicolumn{1}{l|}{\hlc{LightGreen2}{0.084}} & \multicolumn{1}{l|}{\hlc{LightGreen3}{0.171}} & \hlc{LightGreen2}{0.178} \\ \hline
\multicolumn{1}{|l|}{Bangla-T5} & \multicolumn{1}{l|}{\hlc{LightGreen3}{0.078}} & \multicolumn{1}{l|}{\hlc{LightGreen1}{0.084}} & \multicolumn{1}{l|}{\hlc{LightGreen3}{0.063}} & \multicolumn{1}{l|}{\hlc{LightGreen2}{0.068}} & \multicolumn{1}{l|}{\hlc{LightGreen3}{0.138}} & \hlc{LightGreen1}{0.155} \\ \hline
\multicolumn{1}{|l|}{BLOOM} & \multicolumn{1}{l|}{\hlc{LightGreen3}{0.058}} & \multicolumn{1}{l|}{\hlc{LightGreen1}{0.084}} & \multicolumn{1}{l|}{\hlc{LightGreen3}{0.054}} & \multicolumn{1}{l|}{\hlc{LightGreen1}{0.073}} & \multicolumn{1}{l|}{\hlc{LightGreen3}{0.153}} & \hlc{LightGreen1}{0.167} \\ \hline
\multicolumn{7}{|c|}{\textbf{Hindi -> Bengali}} \\ \hline
\multicolumn{1}{|l|}{GPT2-BN} & \multicolumn{1}{l|}{\hlc{LightGreen3}{0.027}} & \multicolumn{1}{l|}{\hlc{LightGreen2}{0.030}} & \multicolumn{1}{l|}{0.064} & \multicolumn{1}{l|}{\hlc{LightGreen1}{0.073}} & \multicolumn{1}{l|}{\hlc{LightGreen3}{0.140}} & \hlc{LightGreen3}{0.139} \\ \hline
\multicolumn{1}{|l|}{mT5-base} & \multicolumn{1}{l|}{\hlc{LightGreen3}{0.102}} & \multicolumn{1}{l|}{\hlc{LightGreen1}{0.116}} & \multicolumn{1}{l|}{\hlc{LightGreen3}{0.076}} & \multicolumn{1}{l|}{\hlc{LightGreen1}{0.087}} & \multicolumn{1}{l|}{\hlc{LightGreen3}{0.162}} & \hlc{LightGreen1}{0.177} \\ \hline
\multicolumn{1}{|l|}{Bangla-T5} & \multicolumn{1}{l|}{\hlc{LightGreen3}{0.096}} & \multicolumn{1}{l|}{\hlc{LightGreen1}{0.103}} & \multicolumn{1}{l|}{\hlc{LightGreen3}{0.081}} & \multicolumn{1}{l|}{\hlc{LightGreen1}{0.088}} & \multicolumn{1}{l|}{\hlc{LightGreen3}{0.161}} & \hlc{LightGreen1}{0.174} \\ \hline
\multicolumn{1}{|l|}{BLOOM} & \multicolumn{1}{l|}{\hlc{LightGreen3}{0.069}} & \multicolumn{1}{l|}{\hlc{LightGreen3}{0.069}} & \multicolumn{1}{l|}{\hlc{LightGreen3}{0.044}} & \multicolumn{1}{l|}{\hlc{LightGreen3}{0.045}} & \multicolumn{1}{l|}{\hlc{LightGreen3}{0.103}} & \hlc{LightGreen3}{0.104} \\ \hline
\end{tabular}
\caption{\footnotesize Few-shot results of the fine-tuned models for the synthetic transfer of EN $\rightarrow$ BN \& HI $\rightarrow$ BN. \hlc{LightGreen1}{Green} denotes performance gain (darker denotes larger gain) with respect to STx0 (see Appendix \ref{synthApp} for EN $\rightarrow$ HI \& BN $\rightarrow$ HI).}
\label{tab:fewshotSynth}
\end{table}

\noindent For the Hindi language, the mT5-base model exhibits the highest BLEU (\textbf{B-2}: 0.175, \textbf{B-3}: 0.123) while the BLOOM model achieves the second highest score in BLEU (\textbf{B-2}: 0.145,\textbf{ B-3}: 0.108) score. ChatGPT demonstrates the highest performance in terms of METEOR (0.166) score and ROUGE-1 (0.261) score.
Regarding \textbf{BERTScore}, the  mT5-base achieves the highest score (0.715) followed by BLOOM with the second-highest score (0.712).
Similar to the Bengali language, we also observe that BLOOM achieves the lowest performance in terms of \textbf{diversity} (0.064) and \textbf{novelty} (0.637). In contrast, similar to Bengali, ChatGPT demonstrates the highest performance, while \texttt{GPT2-HI} exhibits the second-highest score. While we observe that ChatGPT achieves higher scores in diversity and novelty for both languages, this is primarily due to the model generating longer responses with diverse and sometimes irrelevant tokens thus resulting high scores. However, when evaluated based on the BLEU score, the fine-tuned models (Bangla-T5, mT5-base, BLOOM, etc.) consistently outperform the ChatGPT model (refer to Appendix \ref{moreExample} for examples).
 When considering non-abusiveness, BLOOM and mT5-base achieve good scores. However, GPT2-HI and \texttt{docT5query-Hindi} achieve lower scores, indicating that these models often generate abusive speech.
In terms of human judgments, we observe that the BLOOM model achieves the highest score in all metrics, while the mT5-base demonstrates the second-highest performance. Similar to Bengali, ChatGPT exhibits poor performance in terms of the \textbf{choose-or-not} metric. Our rationale for including ChatGPT was to investigate the performance of a large language model (in terms of the number of parameters) in a zero-shot setting. The objective was to assess whether such a model could perform at par with fine-tuned smaller models. Our observations have highlighted the inherent value of fine-tuning, especially for low-resource languages like Bengali and Hindi.

Overall, these large language models can generate CSs for low-resource languages. However, the BLOOM model generates less diverse and repetitive counterspeeches in response to abusive speech.

\begin{table*}
\centering
\includegraphics[width=0.75\textwidth]{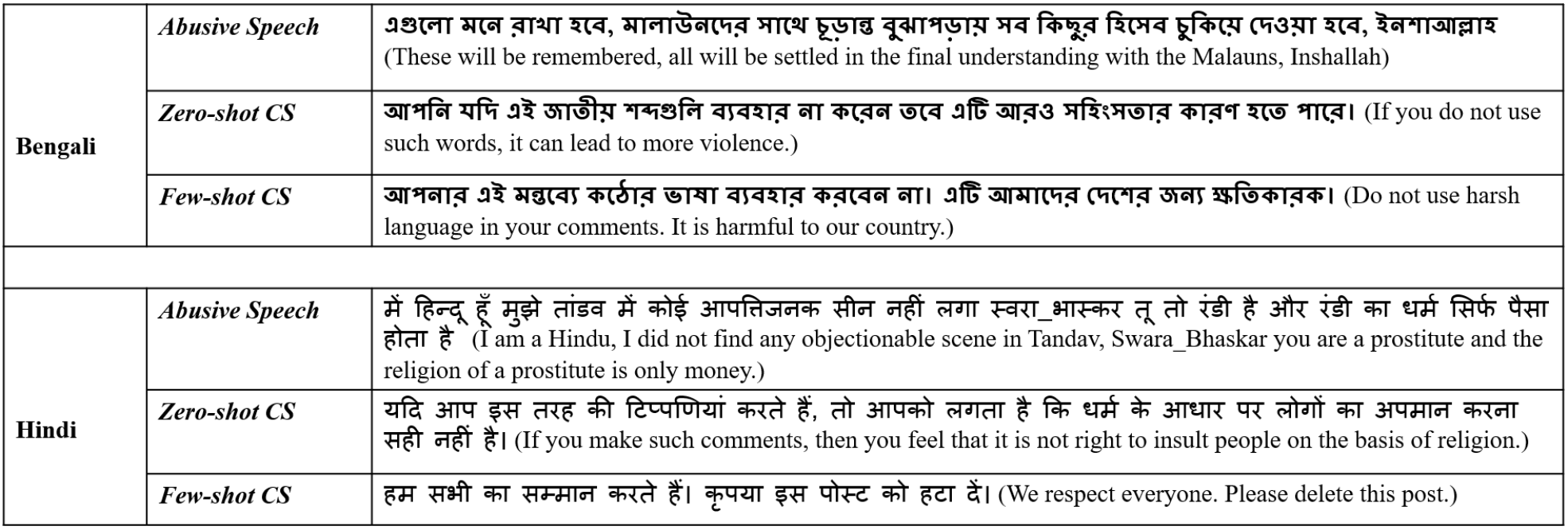}
 \caption{\footnotesize Examples of AS-CS pairs generated by the mT5-base model in zero-shot \& few-shot setting(\textbf{STx2}) for \textbf{HI $\rightarrow$ BN} \& \textbf{BN $\rightarrow$ HI} synthetic transfer. In zero-shot, no gold-label AS-CS pairs were used for training the model.}
\label{zero-shot}
\end{table*}

\subsection{Performance of the joint training}
For this experiment, we focus on the mT5-base and BLOOM models due to their capability to handle both Bengali and Hindi languages together. In Table \ref{tab2f}, we show the performance of joint training. We see that mT5-base achieves the highest BLEU and METEOR scores for both Bengali and Hindi languages. Similar to the monolingual setting, the BLOOM model exhibits low \textbf{diversity} score, indicating that the BLOOM model generates repetitive responses. In terms of human judgment,both models receive high scores for \textbf{grammaticality} (GRE) in both Bengali and Hindi, implying their production of grammatically correct responses. However, the \textbf{specificity} (SPE) score is less than three for both the models for Bengali and for the BLOOM model for Hindi, indicating that these models produce more generalized responses.

In conclusion, joint training can be employed if a generalizable model is desired to generate counterspeeches for multiple languages.

\subsection{Performance of the synthetic transfer}
In Table \ref{tab3} \&  \ref{tab4}, we show the performance of the \textbf{STx0} where we synthetically generate AS-CS pairs from the existing dataset. As expected, the performances are less compared to the monolingual setting for both languages. Table \ref{tab3} reveals that for the Bengali test set, the models trained with \textbf{HI $\rightarrow$ BN} translated synthetic dataset achieve better scores compared to the \textbf{EN $\rightarrow$ BN} translated synthetic dataset. The human evaluation further shows that the generated counterspeeches are of inferior quality for the models trained with \textbf{EN $\rightarrow$ BN}  translated synthetic dataset.
 Similarly, in Table \ref{tab4},  we observe that for the Hindi test set, the models trained with \textbf{BN $\rightarrow$ HI} translated synthetic dataset achieve better scores compared to the \textbf{EN $\rightarrow$ HI}  translated synthetic dataset. Human evaluation also indicates an inferior generation of counterspeeches for the models trained with \textbf{EN $\rightarrow$ HI}  translated synthetic dataset. Among the models trained with \textbf{BN $\rightarrow$ HI}  translated dataset, we observe \texttt{docT5Query-Hindi} and mT5-base models generate counterspeeches with higher scores for human evaluation metrics; however, GPT2-HI and BLOOM show poor performance.

In summary, synthetic transfer schemes work better between Bengali and Hindi languages. This may be attributed by their membership in the  \href{https://en.wikipedia.org/wiki/Indo-Aryan_languages}{Indo-Aryan language family} and the socio-linguistic dissimilarity of English from Hindi and Bengali. One key consideration that motivated our approach is that English datasets are predominantly shaped by Western cultural contexts, which may not directly align with the cultural nuances of Hindi and Bengali. This cultural misalignment could indeed impact the effectiveness of translations. Our experiment aimed to underscore the enhanced transferability between two closely related languages, emphasizing the shared linguistic structure corresponding to \textit{subject} $\rightarrow$ \textit{object} $\rightarrow$ \textit{verb} order in both Bengali and Hindi sentences, as opposed to \textit{subject} $\rightarrow$ \textit{verb} $\rightarrow$ \textit{object} order in English sentences. Table \ref{tab:fewshotSynth} shows the few-shot performance of the synthetic transfer where we add the actual gold AS-CS pairs to fine-tune the models further. Overall we observe adding gold AS-CS gives steady improvements in terms of different overlapping metrics. Hence we recommend instead of developing datasets from scratch, one can use the existing annotated datasets to establish the initial models by performing the synthetic transfer and then fine-tune it for the target language using a small set of gold instances. Table \ref{zero-shot} shows some counterspeeches generated in zero-shot \& few-shot settings. 
For the Bengali CS generation, in zero-shot setting, we observe that the CS supports the AS by saying \textit{``if you do not use such words, it can lead to more violence''}\footnote{Translated to English.} -- ideally, it should have been the opposite. The generated CS became pertinent in the few-shot setting as it said, \textit{``do not use harsh language in your comments, it is harmful to our country''} -- the CS indeed argues that the presence of the offensive word `Malaun'\footnote{An offensive word for Hindus.} is harsh and harmful. This shows that the CS generated after the few-shot training is more relevant/semantically consistent.

In summary, no single model shows consistent performance across all settings for both languages. These variations can be attributed to factors such as model architecture, training data, pre-training strategy, hyperparameters, etc. \citet{cai2022compare} also made a similar observation in a low-resource dataset settings.

\setlength{\tabcolsep}{2pt}
\begin{table}[!h]
\centering
\scriptsize
\begin{tabular}{|cccccccc|}
\hline
\multicolumn{4}{|c|}{\textbf{Bengali}} & \multicolumn{4}{c|}{\textbf{Hindi}}  \\ \hline

\multicolumn{1}{|l|}{\textbf{Model}} & \multicolumn{1}{c|}{\textbf{G}} & \multicolumn{1}{c|}{\textbf{E}} & \multicolumn{1}{c|}{\textbf{TER} $\downarrow$} & \multicolumn{1}{l|}{\textbf{Model}} & \multicolumn{1}{c|}{\textbf{G}} & \multicolumn{1}{c|}{\textbf{E}} & \multicolumn{1}{c|}{\textbf{TER} $\downarrow$}\\ \hline

\multicolumn{1}{|l|}{GPT2-BN} & \multicolumn{1}{c|}{{40.56}} & \multicolumn{1}{c|}{{37.56}} & \multicolumn{1}{c|}{{0.0116}} & \multicolumn{1}{l|}{{GPT2-HI}} & \multicolumn{1}{c|}{{56.63}} & \multicolumn{1}{c|}{{51.39}} & \multicolumn{1}{c|}{{0.0264}}\\ \hline

\multicolumn{1}{|l|}{mT5-base} & \multicolumn{1}{c|}{{13.89}} & \multicolumn{1}{c|}{{12.98}} & \multicolumn{1}{c|}{{0.0031}} & \multicolumn{1}{l|}{{mT5-base}} & \multicolumn{1}{c|}{{22.12}} & \multicolumn{1}{c|}{{21.04}} & \multicolumn{1}{c|}{{0.0044}} \\ \hline

\multicolumn{1}{|l|}{BanglaT5} & \multicolumn{1}{c|}{{18.11}} & \multicolumn{1}{c|}{{17.62}} & \multicolumn{1}{c|}{{0.0019}}& \multicolumn{1}{l|}{{docT5Query}} & \multicolumn{1}{c|}{{22.15}} & \multicolumn{1}{c|}{{21.69}} & \multicolumn{1}{c|}{{0.0006}} \\ \hline

\multicolumn{1}{|l|}{BLOOM} & \multicolumn{1}{c|}{{27.68}} & \multicolumn{1}{c|}{{25.66}} & \multicolumn{1}{c|}{{0.0082}} & \multicolumn{1}{l|}{{BLOOM}} & \multicolumn{1}{c|}{{17.67}} & \multicolumn{1}{c|}{{16.94}} & \multicolumn{1}{c|}{{0.0013}} \\ \hline

\multicolumn{1}{|l|}{ChatGPT} & \multicolumn{1}{c|}{{65.13}} & \multicolumn{1}{c|}{{58.15}} & \multicolumn{1}{c|}{{0.0248}} & \multicolumn{1}{l|}{{ChatGPT}} & \multicolumn{1}{c|}{{103.59}} & \multicolumn{1}{c|}{{60.59}} & \multicolumn{1}{c|}{{0.0350}} \\ \hline

%\multicolumn{1}{|l|}{Mean} & \multicolumn{1}{c|}{{-}} & \multicolumn{1}{c|}{{-}} & \multicolumn{1}{c|}{{0.00992}} & \multicolumn{1}{l|}{{Mean}} & \multicolumn{1}{c|}{{-}} & \multicolumn{1}{c|}{{-}} & \multicolumn{1}{c|}{{0.01354}} \\ \hline

\end{tabular}
\caption{\footnotesize Average length of the generated CS (G) \& edited CS (E) and their TER scores across models.}
\label{tab:TER_mono}
\end{table}

\section{Post-editing evaluation}

We further wanted to assess the utility of the automatically generated responses for the potential moderators who would be using the generated CSs in combating abusive speech on social media. The ideal case would be if they are needed to make absolutely no changes in the generated CSs before posting them on social media. The larger the number of edits they would need to make in the generated CS, the lesser would be its utility. We therefore asked human judges to make necessary edits they would perform before posting the responses on social media. This experiment focused on CS generated in the monolingual setting. We used the translation edit rate (TER)~\cite{snover-etal-2006-study}, a metric analogous to the edit distance to quantify the dissimilarity between the generated CS and edited CS.
This experiment exclusively considers posts selected during human evaluation (CHO=1), calculating TER and the average length of the counterspeech. The results are noted in Table \ref{tab:TER_mono}.

An observation across all models indicates that ChatGPT-generated CSs tend to be lengthy. Hence, annotators had to eliminate certain portions of unnecessary text during the editing process, resulting in a higher TER for ChatGPT in both languages. The average length of generated CS is $\sim$65 for Bengali and $\sim$103 for Hindi. We believe longer CSs can be cumbersome to read and have minimal impact on the abusive speaker. In contrast, BLOOM and mT5-based models exhibit a relatively lower average length of CS, making them more suitable for mitigating abusive speech.

%Further, we calculated the mean TER across all models. For Bengali, the mean TER is 0.0099, and for Hindi, it is 0.0135. This implies that, on average, 1 out of 108 words needs to be edited to generate a human-like CS for Bengali, and 1 out of 75 words needs editing for Hindi.

\section{Conclusion}
Counterspeech generation using neural architecture-based language models has started gaining attention for interventions against hostility. This paper presents the first attempt at CS generation for the Bengali and Hindi languages,  investigating several generation models. To facilitate this, we create a new benchmark dataset of  5,062 AS-CS pairs, of which 2,460 pairs are in Bengali and 2,602 pairs are in Hindi. We experiments with several interlingual transfer mechanisms. Our findings indicate that the overall monolingual setting exhibits the best performance across all the setups. Joint training can be performed if one omnipresent model is  beneficial to generate CSs for multiple languages. We also notice synthetic transferability  yields better results when languages belong to the same language family.

In future, we plan to explore methods for improving specificity by using various types of knowledge (e.g., facts, events, and named entities) from external resources. Further, we plan to add controllable parameters to the counterspeech generation setup, enabling moderators to customize the counterspeech toward a specific technique we have discussed.

\section*{Limitations}
\label{Limit}
There are a few limitations of our work. First, we have focused solely on generating counterspeech for Bengali and Hindi. Further experimentation should be conducted to address the problem of counterspeech generation in other low-resource languages. By expanding our research to include a broader range of languages, we can better understand the challenges and opportunities in generating effective counterspeech across diverse linguistic contexts. Second, we did not incorporate external knowledge, resources, or facts to enhance the generation of counterspeech. Utilizing such additional information could improve counterspeech generation performance by providing more context and accuracy. Furthermore, while we aim to introduce controllable parameters to customize counterspeech, there are challenges in determining the optimal settings for these parameters. Striking the right balance between customization and maintaining ethical boundaries requires careful consideration and further research.

\section*{Ethics Statement}
\subsection{User privacy}
Although our database comprises actual abusive speeches crawled from Twitter, we do not include any personally identifiable information about any user. We follow standard ethical guidelines~\cite{rivers-ethical}, not making any attempts to track users across sites or deanonymize them.
\subsection{Biases}
Any biases noticed in the dataset are unintended, and we have no desire to harm anyone or any group.
\subsection{Potential harms of CS generation models}
Although we observe that these large language models can generate counterspeeches, it is still very far from being coherent and meaningful across the board~\cite{bender2021dangers}. Hence, we do not endorse the deployment of fully automatic pipelines for countering abusive speech~\cite{de2021toxicbot}. Instead, it can be useful as a helping hand to counter speakers in drafting responses to abusive speech.
\subsection{Intended use}
We share our data to encourage more research on low-resource counterspeech generation. We only release the dataset for research purposes and neither grant a license for commercial use nor for malicious purposes.

\bibliography{anthology}
\bibliographystyle{acl_natbib}

\appendix

\section{Annotation guidelines}
\label{sec:anngd}
\subsection{Motivation}
Toxic language is prevalent in online social media platforms, presenting a significant challenge. While methods like user bans or message deletion exist, they can potentially infringe upon the principle of free speech. In this task, our objective is to propose a solution that generates counter-speech in response to abusive language, fostering a more constructive online discourse.

\subsection{Task}
In order to effectively combat abusive language, your task is to craft a well-constructed counter-speech using the recommended strategies outlined in the annotation guidelines. Please ensure that the generated response is clearly marked as a counter-speech, and don't forget to annotate the specific strategy employed to generate the counter-speech. This approach will help us analyze and evaluate the effectiveness of various strategies in addressing abusive language.

\subsection{Recommended strategies}
There could be several techniques to counter abusive speech. \citet{benesch2016considerations} distinguish eight such strategies that counter speakers typically use. However, not all strategies help to reduce the propagation of abusive speech. Therefore the author further recommended strategies that can be beneficial to develop positive influence. We discuss these recommended strategies below.
\begin{itemize}
    \item \textit{\textbf{Warning of consequences}} (WoC): In this strategy, the counter speakers often warn of the possible consequences of posting hateful content on public platforms like Twitter. This can occasionally drive the original speaker of the abusive speech to delete his/her source post.
    \item \textit{\textbf{Pointing out hypocrisy}}: In this strategy, the counter speaker points out the hypocrisy or contradiction in the user's (abusive) statements. In order to discredit the accusation, the individual may illustrate and rationalize their previous behavior, or if they are persuadable, resolve to evade the dissonant behavior in the future.
    \item \textit{\textbf{Shaming and labeling}}: In this strategy, the counter speaker denounces the post as disgusting, abusive, racist, bigoted, misogynistic, etc. This strategy can help the counter speakers reduce the hateful post's impact.
    \item \textit{\textbf{Affiliation}}: Affiliation is ``... establishing,  maintaining, or restoring a positive affective relationship with another person or group''. People  are  more likely to credit the counterspeech of those with whom they affiliate since they  tend  to ``evaluate  ingroup  members  as more  trustworthy,  honest,  loyal,  cooperative,  and  valuable to the group than outgroup members''.
    \item \textit{\textbf{Empathy}}: In this strategy, the counter speaker uses an empathetic, kind, peaceful tone in response to hateful messages to undermine the abusive post. Changing the tone of a hateful conversation is an effective way of ending the exchange. Although we have little evidence that this will change behavior in the long term, it may prevent the rise of hate speech used at the present moment.
    \item \textit{\textbf{Humor and sarcasm}}: Humor is one of the most effective tools used by counter speakers to combat hostile speech. It can de-escalate conflicts and can be used to garner more attention toward the topic. Humor in online environments also eases execration, supports other online speakers, and facilitates social cohesion.
\end{itemize}

\subsection{Dealing with post-annotation stress}

We gave the following piece of advice to our annotators -- ``We understand that the task at hand is challenging and may have an emotional impact on you. It is important to prioritize your well-being while undertaking these annotations. We strongly recommend taking regular breaks throughout the process. If you find yourself experiencing any form of stress or difficulty, please reach out to the mentors for support. They are there to assist you and may advise you to pause the annotations for a period of 2-3 days to ensure your well-being.

In addition, there is a helpful resource available for you to manage stress in any challenging situation. Please visit \url{https://yourdost.com/} for support and guidance.

We would also wish to provide you with some pointers on dealing with moderator stress. You can find important insights at \citet{Two_Hat_2020}. In addition, please reach out to your mentors for additional support.

We sincerely appreciate your participation in this annotation task. Your contribution is crucial in furthering our understanding of such societal issues.''

\section{Implementation details}
\label{imple_det}
All the models are coded in Python, using the Pytorch library. All training and evaluation have been performed on a Tesla P100-PCIE (16GB) machine with differing batch sizes (GPT2-HI: 1, GPT-BN: 1, mT5-base: 4, \texttt{docT5Query-Hindi}: 4, \texttt{BanglaT5}: 8, BLOOM: 4) depending on the model architecture. All the models were run up to 50 epochs with  Adafactor optimizer~\cite{shazeer2018adafactor} having a learning rate of $2e-5$. We save the models for the best validation perplexity score~\cite{zhang2019dialogpt}. We also use EarlyStopping patience when validation perplexity decreases by less than $1e-4$. For ChatGPT, we utilized the \texttt{gpt-3.5-turbo} model, a chatbot based on the GPT-3.5 language model. The ``temperature'' parameter was set to 0 to minimize variations in ChatGPT-generated outputs. When generating responses, the ``max\_tokens'' parameter was set to 300.

\begin{table}[h]
\scriptsize
\centering
\begin{tabular}{|lllllll|}
\hline
\multicolumn{7}{|c|}{\textbf{English->Hindi}} \\ \hline
\multicolumn{1}{|l|}{} & \multicolumn{2}{c|}{\textbf{B-2}} & \multicolumn{2}{c|}{\textbf{M}} & \multicolumn{2}{c|}{\textbf{ROU}} \\ \hline
\multicolumn{1}{|l|}{Model} & \multicolumn{1}{l|}{STx1} & \multicolumn{1}{l|}{STx2} & \multicolumn{1}{l|}{STx1} & \multicolumn{1}{l|}{STx2} & \multicolumn{1}{l|}{STx1} & STx2 \\ \hline
\multicolumn{1}{|l|}{GPT2-HI} & \multicolumn{1}{l|}{\hlc{LightGreen3}{0.088}} & \multicolumn{1}{l|}{\hlc{LightGreen3}{0.088}} & \multicolumn{1}{l|}{\hlc{LightGreen3}{0.132}} & \multicolumn{1}{l|}{\hlc{LightGreen3}{0.131}} & \multicolumn{1}{l|}{\hlc{LightGreen2}{0.239}} & \hlc{LightGreen3}{0.231} \\ \hline
\multicolumn{1}{|l|}{mT5-base} & \multicolumn{1}{l|}{\hlc{LightGreen3}{0.156}} & \multicolumn{1}{l|}{\hlc{LightGreen2}{0.161}} & \multicolumn{1}{l|}{\hlc{LightGreen3}{0.115}} & \multicolumn{1}{l|}{\hlc{LightGreen3}{0.117}} & \multicolumn{1}{l|}{\hlc{LightGreen3}{0.226}} & \hlc{LightGreen3}{0.227} \\ \hline
\multicolumn{1}{|l|}{docT5Query} & \multicolumn{1}{l|}{\hlc{LightGreen3}{0.142}} & \multicolumn{1}{l|}{\hlc{LightGreen2}{0.146}} & \multicolumn{1}{l|}{\hlc{LightGreen3}{0.106}} & \multicolumn{1}{l|}{\hlc{LightGreen2}{0.111}} & \multicolumn{1}{l|}{\hlc{LightGreen3}{0.216}} & \hlc{LightGreen2}{0.219} \\ \hline
\multicolumn{1}{|l|}{BLOOM} & \multicolumn{1}{l|}{0.111} & \multicolumn{1}{l|}{\hlc{LightGreen1}{0.127}} & \multicolumn{1}{l|}{0.087} & \multicolumn{1}{l|}{\hlc{LightGreen3}{0.096}} & \multicolumn{1}{l|}{0.197} & \hlc{LightGreen3}{0.210} \\ \hline
\multicolumn{7}{|c|}{\textbf{Bengali->Hindi}} \\ \hline
\multicolumn{1}{|l|}{GPT2-HI} & \multicolumn{1}{l|}{\hlc{LightGreen3}{0.090}} & \multicolumn{1}{l|}{\hlc{LightGreen3}{0.089}} & \multicolumn{1}{l|}{\hlc{LightGreen3}{0.138}} & \multicolumn{1}{l|}{\hlc{LightGreen3}{0.136}} & \multicolumn{1}{l|}{0.247} & 0.238 \\ \hline
\multicolumn{1}{|l|}{mT5-base} & \multicolumn{1}{l|}{0.165} & \multicolumn{1}{l|}{0.168} & \multicolumn{1}{l|}{\hlc{LightGreen3}{0.123}} & \multicolumn{1}{l|}{\hlc{LightGreen3}{0.126}} & \multicolumn{1}{l|}{\hlc{LightGreen3}{0.229}} & \hlc{LightGreen2}{0.235} \\ \hline
\multicolumn{1}{|l|}{docT5Query} & \multicolumn{1}{l|}{\hlc{LightGreen3}{0.148}} & \multicolumn{1}{l|}{\hlc{LightGreen2}{0.154}} & \multicolumn{1}{l|}{\hlc{LightGreen3}{0.106}} & \multicolumn{1}{l|}{\hlc{LightGreen1}{0.114}} & \multicolumn{1}{l|}{\hlc{LightGreen3}{0.203}} & \hlc{LightGreen1}{0.214} \\ \hline
\multicolumn{1}{|l|}{BLOOM} & \multicolumn{1}{l|}{0.092} & \multicolumn{1}{l|}{0.095} & \multicolumn{1}{l|}{0.062} & \multicolumn{1}{l|}{0.065} & \multicolumn{1}{l|}{0.147} & 0.155 \\ \hline
\end{tabular}
\caption{Few-shot results of the fine-tuned models for the synthetic transfer of EN $\rightarrow$ HI \& BN $\rightarrow$ HI. \hlc{LightGreen1}{Green} denotes performance gain (darker denotes larger gain) with respect to STx0.}
\end{table}

\section{Synthetic transfer performance}
\label{synthApp}

In Table \ref{tab:fewshotSynth}, we show the few-shot performance of the synthetic transfer for the EN->HI and HI $\rightarrow$ BN settings, where we add the actual gold AS-CS pairs to fine-tune the models further.

\begin{table*}
\centering
\begin{minipage}[t]{.45\textwidth}
  \centering
		\includegraphics[width=\textwidth]{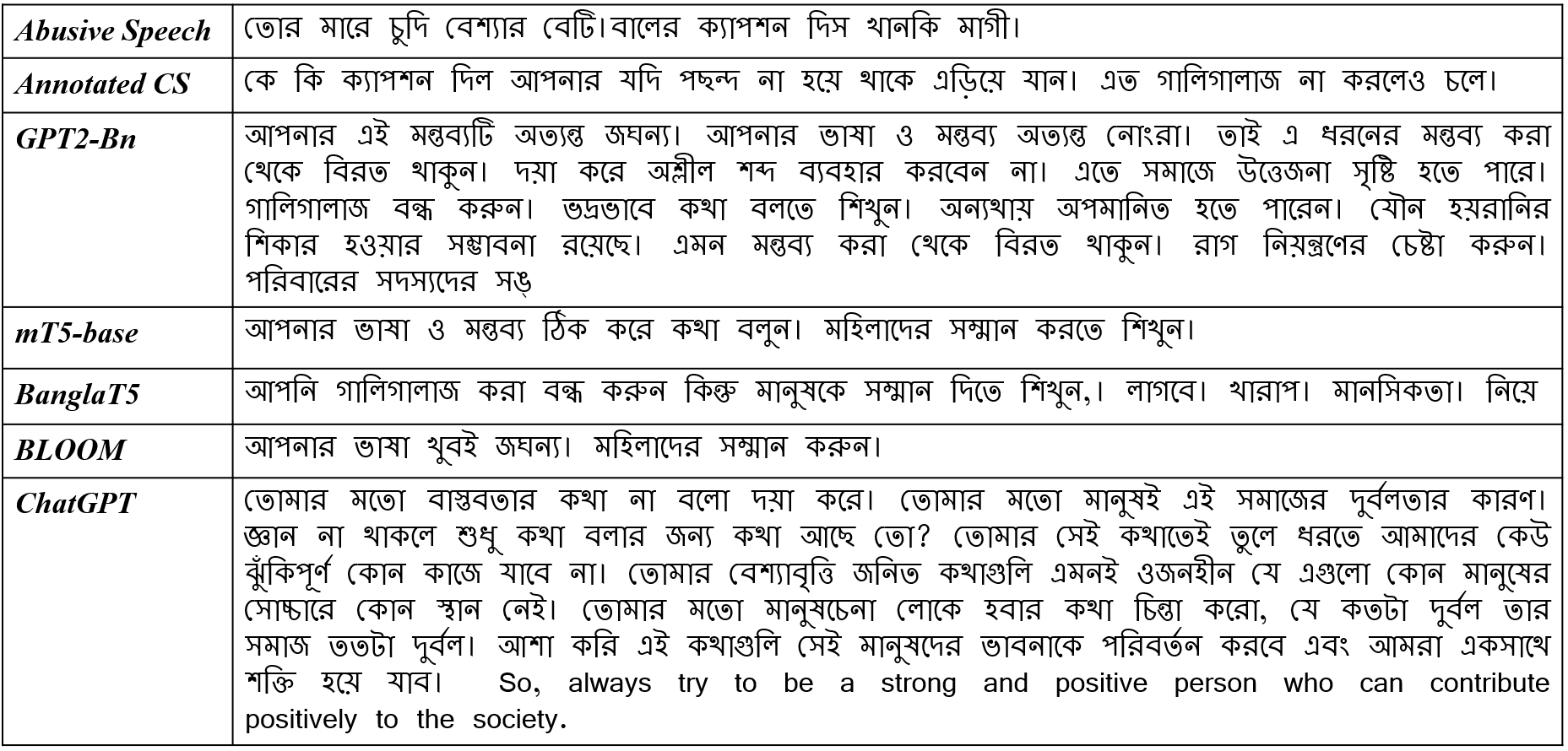}
    \caption{Examples of Bengali AS-CS pairs generated by the models.}
	\label{tab:bengaliCSEx}
\end{minipage}%
\quad
\begin{minipage}[t]{.45\textwidth}
  \centering
  \includegraphics[width=\textwidth]{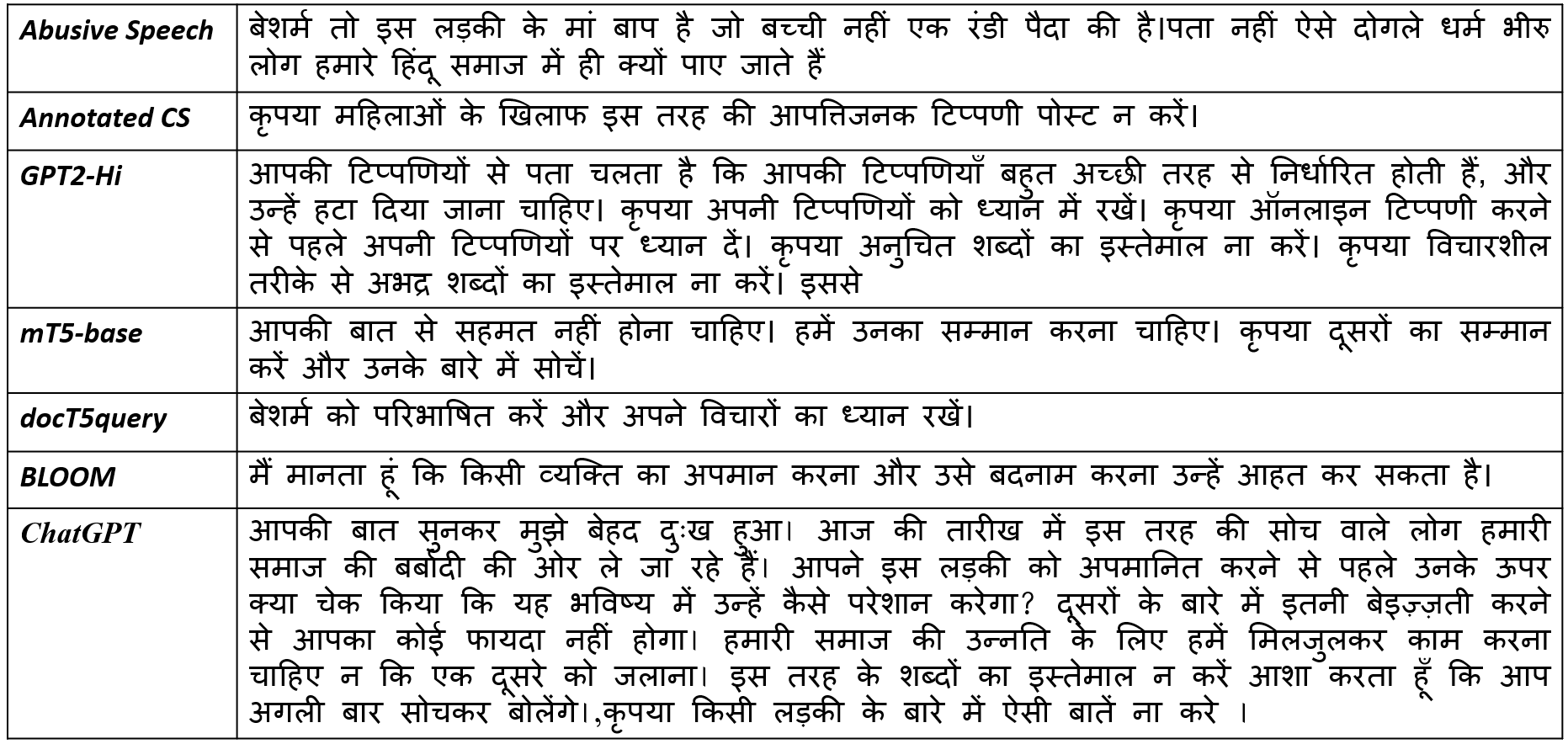}
    \caption{Examples of Hindi AS-CS pairs generated by the models.}
  \label{tab:hindiCSEx}
\end{minipage}
\end{table*}

\section{More examples}
\label{moreExample}
In Tables \ref{tab:bengaliCSEx} and \ref{tab:hindiCSEx}, we present additional examples of the generated CS in the monolingual setting. As observed, the responses generated by ChatGPT are longer compared to those of the other models. In, the generated CSs are not always perfect; hence, more research should be conducted to improve the CS generation of these models.

\section{Generated counterspeech type}
\label{genCSType}

We also conducted an analysis to observe the types of CS being generated. In Figure \ref{fig:strat}, we present the distribution of different types of CS of the annotated data. We expect the models to learn these types of CS during fine-tuning. The experiment was conducted on CSs generated in a monolingual setting and for the counterspeech selected (CHO=1) during manual evaluation. In Table \ref{tab:CS_cls}, we show the types of CSs generated by the different models. We observe, in general, that most of the CSs are classified as \textit{warning of consequences} (WOC), \textit{shaming and labeling}, and \textit{empathy}. However, not all models exhibit the same distribution, and almost all models struggle to generate CS of types \textit{pointing out hypocrisy}, \textit{affiliation}, and \textit{humor and sarcasm}. While this study was conducted with a limited number of generated CS, a more in-depth analysis is required for a comprehensive understanding and type-suitable generation of CSs.

\begin{table}[h]
\scriptsize
\centering
\begin{tabular}{|lllllll|}
\hline
\multicolumn{7}{|c|}{\textbf{Bengali}}                                                                                                                                                                                                    \\ \hline
\multicolumn{1}{|l|}{\textbf{Model}} & \multicolumn{1}{l|}{\textbf{WOC}} & \multicolumn{1}{l|}{\textbf{S\&L}} & \multicolumn{1}{l|}{\textbf{EMP}} & \multicolumn{1}{l|}{\textbf{POH}} & \multicolumn{1}{l|}{\textbf{AFF}} & \textbf{H\&S} \\ \hline
\multicolumn{1}{|l|}{GPT2-BN}        & \multicolumn{1}{l|}{0.27}         & \multicolumn{1}{l|}{0.16}          & \multicolumn{1}{l|}{0.00}         & \multicolumn{1}{l|}{0.00}         & \multicolumn{1}{l|}{0.00}         & 1.00          \\ \hline
\multicolumn{1}{|l|}{mT5-base}       & \multicolumn{1}{l|}{0.12}         & \multicolumn{1}{l|}{0.31}          & \multicolumn{1}{l|}{0.11}         & \multicolumn{1}{l|}{0.40}         & \multicolumn{1}{l|}{0.00}         & 0.00          \\ \hline
\multicolumn{1}{|l|}{BanglaT5}       & \multicolumn{1}{l|}{0.14}         & \multicolumn{1}{l|}{0.31}          & \multicolumn{1}{l|}{0.10}         & \multicolumn{1}{l|}{0.20}         & \multicolumn{1}{l|}{0.00}         & 0.00          \\ \hline
\multicolumn{1}{|l|}{BLOOM}          & \multicolumn{1}{l|}{0.39}         & \multicolumn{1}{l|}{0.08}          & \multicolumn{1}{l|}{0.30}         & \multicolumn{1}{l|}{0.00}         & \multicolumn{1}{l|}{0.40}         & 0.00          \\ \hline
\multicolumn{1}{|l|}{ChatGPT}        & \multicolumn{1}{l|}{0.08}         & \multicolumn{1}{l|}{0.14}          & \multicolumn{1}{l|}{0.49}         & \multicolumn{1}{l|}{0.40}         & \multicolumn{1}{l|}{0.60}         & 0.00          \\ \hline
\multicolumn{7}{|c|}{\textbf{Hindi}}                                                                                                                                                                                                      \\ \hline
\multicolumn{1}{|l|}{GPT2-HI}        & \multicolumn{1}{l|}{0.22}         & \multicolumn{1}{l|}{0.20}          & \multicolumn{1}{l|}{0.06}         & \multicolumn{1}{l|}{0.00}         & \multicolumn{1}{l|}{0.14}         & 0.00          \\ \hline
\multicolumn{1}{|l|}{mT5-base}       & \multicolumn{1}{l|}{0.13}         & \multicolumn{1}{l|}{0.24}          & \multicolumn{1}{l|}{0.22}         & \multicolumn{1}{l|}{0.00}         & \multicolumn{1}{l|}{0.33}         & 1.00          \\ \hline
\multicolumn{1}{|l|}{docT5Query}     & \multicolumn{1}{l|}{0.03}         & \multicolumn{1}{l|}{0.24}          & \multicolumn{1}{l|}{0.06}         & \multicolumn{1}{l|}{0.00}         & \multicolumn{1}{l|}{0.20}         & 0.00          \\ \hline
\multicolumn{1}{|l|}{BLOOM}          & \multicolumn{1}{l|}{0.55}         & \multicolumn{1}{l|}{0.03}          & \multicolumn{1}{l|}{0.59}         & \multicolumn{1}{l|}{0.00}         & \multicolumn{1}{l|}{0.00}         & 0.00          \\ \hline
\multicolumn{1}{|l|}{ChatGPT}        & \multicolumn{1}{l|}{0.07}         & \multicolumn{1}{l|}{0.29}          & \multicolumn{1}{l|}{0.07}         & \multicolumn{1}{l|}{0.00}         & \multicolumn{1}{l|}{0.33}         & 0.00          \\ \hline
\end{tabular}
\caption{Different types of counterspeech generated by different models. Values are normalized column-wise between 0 to 1. WOC: \textit{warning of consequences}, S\&L: \textit{shaming and labeling}, EMP: \textit{empathy}, POH: \textit{pointing out hypocrisy}, AFF: \textit{affiliation}, H\&S: \textit{humor and sarcasm}.}
\label{tab:CS_cls}
\end{table}

\end{document}